\documentclass[12pt]{article}

\usepackage{scicite}
\usepackage{graphicx}
\usepackage{hyperref}

\usepackage{times}

\newenvironment{sciabstract}{%
\begin{quote} \bf}
{\end{quote}}

\newcounter{lastnote}

\title{{\it Large-scale Generative Simulation Artificial Intelligence\/}: the Next Hotspot in Generative AI}

\author
{Qi Wang,$^{1}$ Yanghe Feng,$^{2}$ Jincai Huang,$^{2\ast}$ Yiqin Lv,$^{3}$ Zheng Xie,$^{1}$ Xiaoshan Gao$^{4\ast}$\\
\\
\normalsize{$^{1}$Kaiyuan Mathematical Sciences Institute,Changsha, 410000, China}\\
\normalsize{$^{2}$College of Systems Engineering, National University of Defense Technology, Changsha, 410073, China}\\
\normalsize{$^{3}$College of Sciences, National University of Defense Technology, Changsha, 410073, China}\\
\normalsize{$^{4}$Chinese Academy of Sciences, Peking, 100190, China}\\
\\
\normalsize{$^\ast$Correspondence Authors: huangjincai@nudt.edu.cn; xgao@mmrc.iss.ac.cn.}
}

\date{}

\begin{document} 


\baselineskip24pt


\maketitle


\begin{sciabstract}
  
\end{sciabstract}


\section*{Introduction}

Nowadays, big data, deep learning models, optimization methods, and computational power are essential in promoting the development of artificial intelligence. Recent advances are focused on generative artificial intelligence (GenAI), which paves unprecedented paths to exploring the mechanisms behind the creation of new things (texts, images, videos, or other contents) rather than simply performing discriminative learning tasks. 

GenAI's emergence, e.g., the large model, has changed the landscape of deep learning research and inevitably influenced individuals in both work and life. Furthermore, GenAI holds tremendous potential to reshape robotics research, national governance, and life sciences. Consequently, a pressing question arises: "Will GenAI inspire a new round of technological revolution?" In answer to this question, the commentary pursues deeper insights into generative modeling, highlights critical considerations when building future GenAI models, identifies existing dilemmas and nominates the next hotspot for generalizing large models to more practical scenarios. Throughout the commentary, we use $x\in\mathcal{X}$, $y\in\mathcal{Y}$, and $z\in\mathcal{Z}$ to denote the explanatory variable, response variable, and latent variable, respectively. For tasks or operators on datasets $\tau$, we represent them as distributions in the form of $p\left(\tau\right)$.

\section*{GenAI can do more than AIGC}

Now widespread popularity of GenAI models stems from their ability of artificial intelligence generated content (AIGC). Technically, the deep generative model empowers GenAI's numerous capabilities beyond standard AIGC. Among them, we stress three practical ones in \textbf{Fig. (1.a)}: data compression, representation disentanglement, and causal inference.

In the big data era, minimizing the number of bits required to store and transmit information, known as data compression, is crucial. This function is particularly essential in time-sensitive services with memory constraints, such as edge computing. Some generative models, such as the vector quantized variational autoencoder or deep variational information bottleneck models, excel in data compression by finding insufficient statistics of high-dimensional signals. Representation disentanglement refers to the ability to infer statistically independent latent variables that explain different aspects of data generation, e.g., style, color, and pose. It closely relates to the controllable generation, e.g., obtaining samples with only one aspect varied. Causality is also a significant concern in GenAI, and generative models are advantageous in handling high-dimensional variables and discovering structures of causal graphs, allowing for understanding causal effects. Importantly, GenAI with causality enables counterfactual predictions, which renders the potential consequences of a specific intervention that we have yet to execute. For example, with $p\left(y\middle| x,\mathrm{do}\left(z=z_0\right)\right)$, policymakers can evaluate the influence of socio-economic policies, denoted by $z_0$, without incurring additional costs.

Despite these fascinating capabilities, there remain several tricky questions in the field. (i) Is fully representation disentanglement achievable with generative models? (ii) How can we identify causal generative models in the presence of small-scale datasets and many unobserved confounders?

\section*{Experimental design matters in GenAI's adaptability and robustness}
Let us rethink the critical factors contributing to GPT-like models' success. In addition to prompt engineering, the languages’ generative process must be capable of capturing the masked input-output coupling pattern in the corpus, mapping these linked entities to a knowledge base, and continually improving its performance by incorporating new input-output pairs. Hence, when users initiate queries for specific contextual terms, the knowledge base can effectively locate the precise information and provide feedback.

The above process inspires the task distribution design for GenAI. Task diversity nurtures the generalization capability of models across various scenarios, as this aligns with traditional statistical learning theory. The sampled tasks should be representative, covering a broad range of scenarios, particularly in zero-shot or few-shot learning. However, increasing task diversity requires larger model sizes and comes at the cost of higher computational expenses. For instance, GPT-3 has 175 billion parameters and has been trained on over 570GB of text data from diverse tasks. Generating tasks is problem-specific, and masked learning has emerged as one of the most popular heuristics. Nevertheless, exhaustively exploring all scenarios in training large models can be computationally demanding. As an example, consider \textbf{Fig. (1.b)}, where the number of masked scenarios to complete grows exponentially with respect to the complexity of data $\left|\mathcal{X}\right|$ in a combinatorial sense.

Conversely, we raise two key issues: (i) Is there a principle to balance average performance and adaptability to worst-case scenarios, particularly when loss values exhibit heavy tails? (ii) How can we automatically design task distributions in a dataset or instance-wise sense to improve generalization? These issues require further attention in the development of desirable GenAI.

\section*{Geometric priors can be powerful inductive biases in boosting GenAI}
Generally, we refer to constraints or encoded knowledge in hypotheses space as inductive bias. As stated in Max Welling's comment \cite{welling2019} on the Bitter Lesson \cite{sutton2019}, machine learning cannot generalize well without inductive biases. Inductive biases are especially beneficial when dealing with data insufficiency, as it guides the learning process in a more reasonable direction. 

Here we concentrate on the geometric inductive bias, which conserves the geometric structures of datasets. At a high level, these structures resort to symmetry and scale separation principles \cite{bronstein2021geometric}, particularly necessary in generative modeling. Take the equivariance in symmetry as an example: Human cognitive systems can naturally capture the rotation, translation, reflection, and scaling of signals, implying that the reasonable abstraction of concepts is equivariant with respect to these transformations, as shown in \textbf{Fig. (1.c)}. Another way to apply geometric priors is selective data augmentation by imposing transformation in the data space, meaning that data itself can be inductive bias in modeling. Generative modeling transformations of the dataset can also be attractive in deep geometric learning. The geometric prior and recent advances have been verified to be effective in GenAI for scientific discoveries, e.g., molecule design and drug development, better capturing the complex interactions between atoms and predicting the properties of drug candidates. This constitutes a promising avenue for the application of geometric priors in the field of AI4Science.

While geometric priors show promise in GenAI, important questions still need answers: (i) Are there universal routines to automatically generate geometric priors in GenAI applications? (ii) How can we alleviate the computation burden when incorporating them through constraints or data augmentation? Addressing these questions could help us understand the role of geometric priors in generative modeling and facilitate their use in practice.

\subsection*{Multi-views are required to evaluate performance of models for GenAI}
GenAI primarily counts more on the data generation mechanism. Given the inherent subjectivity and variability of specific applications, there exist no universally applicable criteria to evaluate generative performance.

For reliability and usefulness, we propose to establish multi-view evaluation systems that consider fidelity, diversity, and safety in \textbf{Fig. (1.d)}. The fidelity in generation is critical in risk-sensitive applications like dialogue systems in medical science, and standard metrics are log-likelihoods or statistical divergence such as inception score. Diversity is a fundamental characteristic of generative modeling, with the purpose of capturing the complete possible examples in the probability space. At least two factors influence generation diversity: the extent of observability and the complexity of the dataset semantics. Observability extent refers to the level of accessible context details, namely prior information. For example, in image inpainting, the diversity of generated images decreases as more pixels are observed. Empirically in large language models, increasing the corpus’ size and complexity brings more varied and creative text generation. When using the Bayesian framework, efficient probabilistic programming requires stochastic optimization algorithms to avoid posterior or conditional prior collapse to guarantee the diversity \cite{wang2021posterior}. Additionally, security is an increasingly important concern in GenAI. For example, dataset bias, such as deliberate manipulation or unintentional sampling bias, can significantly affect the performance and orientation of large language models. Notably, the trend of GenAI is to allow interactions with open environments, incrementally access Internet information, and evolves in a continual learning way, securing generative models from attacks at a data level seems urgent. Undoubtedly, there is a solid allure for exploring a model-agnostic and domain-agnostic evaluation schema that is end-to-end and integrates multi-views at both the sample and distribution levels.

\section*{Large-scale generative simulation artificial intelligence is on the way}
The concept of GenAI has been developed for decades. Until recently, it has impressed us with substantial breakthroughs in natural language processing and computer vision, actively engaging in industrial scenarios. Noticing the practical challenges, e.g., limited learning resources, and overly dependencies on scientific discovery empiricism, we nominate large-scale generative simulation artificial intelligence (LS-GenAI) as the next hotspot for GenAI to connect. 

The roadmap of GenAI in \textbf{Fig. (1.e)} relies on previously considered elements and can be framed as the doubly generative paradigm for simulation and sequential decision-making. Specifically, the simulation system needs to be identifiable with a few observations, and the decision-making modules can afford fast adaptation utility in time-sensitive scenarios, e.g., autonomous driving. At the intersection of simulation science and artificial intelligence, LS-GenAI also has particular use in robotics and life systems, reducing realistic sampling complexity, accelerating scientific progress, and catalyzing discoveries. One prime example of LS-GenAI 's potential originates from clinical research. In this context, a high-fidelity biomedical simulation system, operating at the individual level, can create environments to allow the examination of the treatment effects on patients and reduce dependencies on expert experience. 

In spite of numerous realistic benefits, developing LS-GenAI is nontrivial. The demands of massive real-world data, the lack of high-fidelity world models, and the weak adaptability of these world models have complicated the process of constructing ubiquitous decision-making systems, e.g., in interventional clinical research. In service of the utilities in LS-GenAI, more sophisticated simulation and learning tools must be integrated. Apart from building high-fidelity simulation environments, or world models5, it is essential to support customization for different decision-making tasks such that the task of our interest can be among them. Meanwhile, the world exhibits a hierarchical structure, such as the atomic, cellular, tissue, and organismal levels in human body systems, or spatiotemporal scales, and retaining multi-scale in generation augmented by symbolic computation can reveal more accurate complex dynamics. Other demands lie in handling partial observability with the inaccessible inherent system state and unpacking the black box to separate function approximation and causal effects. The primary goals of LS-GenAI are to assist in meaningful experimental design and enable fast adaptation of learned skills. Achieving these will ultimately enrich the utilities of GenAI in a broader range of real-world scenarios.

\begin{figure}[ht]
\centering
\includegraphics[height=1.1\textheight]{Figure1.pdf}
\label{fig:stream}
\end{figure}

\bibliography{scibib}

\bibliographystyle{Science}

\section*{Acknowledgements}
We thank Tailin Wu, Xiangming Meng, and Qian Lou for the helpful discussion on large models in an online forum.
This commentary is also in memory of dr. Yanghe Feng's efforts and contributions to developing intelligent decision-making systems in the field.

\end{document}